\documentclass{article}

\PassOptionsToPackage{numbers,compress}{natbib}

\usepackage[dblblindworkshop,final]{neurips_2026}
\workshoptitle{TTCL Workshop 2026}  
\usepackage[utf8]{inputenc}
\usepackage[T1]{fontenc}
\usepackage{hyperref}
\usepackage{url}
\usepackage{booktabs}
\usepackage{amsfonts}
\usepackage{amsmath}
\usepackage{nicefrac}
\usepackage{microtype}
\usepackage{xcolor}
\usepackage{graphicx}
\usepackage{multirow}
\usepackage{subcaption}
\usepackage{enumitem}
\usepackage{tikz}
\usepackage{float}
\usepackage{amsmath}
\usepackage{amssymb}
\usepackage{amsfonts}
\usetikzlibrary{arrows.meta, positioning, shapes.geometric}
\newcommand{\smname}[1]{\texttt{#1}}

\title{DISCERN: Preserve at Write, Resolve at Read for Persistent LLM-Agent Memory}

\author{%
  Chirag Ahuja \quad Vladislav Tanov \quad Ramcharan Kakarla \quad Jonathan Taws \\
  Amazon \\
  \texttt{\{cahuja, vtanov, ramkaka\}@amazon.com}, \texttt{tawsj@amazon.lu} \\
}

\begin{document}
\maketitle

\begin{abstract}

Enterprise agents accumulate memory across dozens of sessions in which facts contradict, get superseded, and blur as similarly named entities collide in embedding space. The field treats this as a \emph{retrieval} problem; we show it is not. On ENT-310, a synthetic enterprise sales-agent memory benchmark of 310 queries (210 answerable, 100 unanswerable), a vanilla Retrieval-Augmented Generation (RAG) reader over a memory store already matches strong memory systems on the recall pillars. Retention and currency are near-saturated, roughly 0.93 to 0.98 for every competent system, so recall is essentially solved. What remains unsolved is \emph{resolution}: which entity a query refers to, which time-state is current, and whether to commit an answer at all or decline when evidence is insufficient. We propose \textbf{DISCERN} (DISambiguation and Cross-session Episodic memoRy Navigation), an agent-memory framework built on a single rule: \emph{preserve at write, resolve at read}. At write time DISCERN builds a strictly additive bi-temporal provenance graph that never merges confusable entities and never drops superseded values. At read time a zero-LLM tool surface drives an agentic loop that navigates this graph and commits exactly one of four gated terminal actions. A four-judge LLM-as-a-Judge council scores two axes: answer correctness on answerable questions, and the answer-versus-abstain decision as a classification problem (precision, recall, F1) on the questions where the right behavior is to abstain. On ENT-310, DISCERN is the best-calibrated system on abstention by a wide margin (F1 0.924 versus at most 0.802 across eight baselines) and leads macro answer correctness under equal pillar weighting (0.861 versus 0.806 for the strongest baseline), a lead driven by entity resolution (0.800, where no baseline exceeds 0.40). Applied with zero tuning to the full LoCoMo~\cite{maharana2024} corpus, the same pipeline again leads macro answer correctness (0.663 versus 0.638) and stays among the best-calibrated on false-premise abstention, so the result is not benchmark-specific.

\end{abstract}


\section{Introduction}
\label{sec:intro}

A seller asking ``where did we land on the prototype for Zephyron?''\footnote{All company names in this paper are fictitious and used for illustration only; any resemblance to real entities is coincidental. The underlying benchmark data (revenues, deal stages, attributes) is synthetic.} expects the agent to recall a decision made three sessions ago. This task is \emph{warm-start}: the answer already lives in prior sessions, and a single context window cannot hold months of an advisor's interactions across dozens of accounts, so that history must be stored and selectively recalled. Persistent memory is therefore a foundational capability for such agents.

A common enterprise design splits this into a write path and a read path~\citep{mem0}. At write time a semantic strategy extracts entity--attribute--value facts (e.g., \emph{Zephyron Cloud Solutions: revenue $=$ \$5.1B}) and a summary strategy produces rolling per-session summaries, both under per-actor namespacing. At read time a retriever pulls the top-$k$ most similar memories ($k{=}10$) from both stores, and one query-enrichment (QE) LLM call resolves which entity the query targets. That call is skipped when the entity is already unambiguous, so the pipeline pays no extra latency on most queries. We take this semantic-plus-summary-plus-enrichment configuration as our strongest baseline.

\textbf{On these benchmarks, recall is largely solved; resolution is not.} On our benchmark this pipeline already retrieves the right evidence: retention and currency are near-saturated (roughly 0.93 to 0.98 for every competent system). Those numbers measure whether the evidence was retrieved, not whether the committed
answer was right, and two problems sit in that gap. \textbf{P1, one-shot disambiguation:} a single
post-retrieval enrichment call resolves ambiguity, with no loop of evidence gathering
and verification. \textbf{P2, no terminal-state contract:} the system always answers,
even when the question is unanswerable from memory, and nothing distinguishes
\textsc{answer}/\textsc{update}/\textsc{clarify}/\textsc{abstain}, so blended, stale,
and hallucinated answers are invisible to its metrics.

We argue the difficulty is not retrieval ranking but a division of labor: a memory system should organize entities, conflicts, and time into an explicit structure ahead of time, and defer only the query-specific commitment to answer time. \textbf{DISCERN} follows one principle, \emph{preserve at write, resolve at read}. Write time does substantial but lossless work: every observation is appended to an immutable per-actor event log, the single source of truth, and eagerly consolidated into a derived spatiotemporal memory graph (an entity-family graph, a bitemporal fact store, and episodes linked to the entities they involve and chained along a session timeline). This consolidation commits to nothing irreversible. Confusable entities are linked but never merged, superseded facts are invalidated softly rather than deleted, and every view regenerates from the log, so the graph is a reversible projection rather than the store of record and the approach stays agnostic to the physical backend. Read time then carries the query-specific resolution: rather than fix a single pipeline, DISCERN treats access as an adaptive navigation policy over the graph and commits to exactly one gated terminal action.

\textbf{Contributions}
\begin{enumerate}[leftmargin=1.6em,itemsep=1pt,topsep=2pt]
    \item \textbf{DISCERN}, a memory system built on \emph{preserve at write, resolve at read}: a strictly additive bi-temporal provenance graph (\S\ref{sec:graph}) plus a read-time loop in which each zero-LLM tool call is a routing decision descending a memory hierarchy toward one of four gated terminal actions (\S\ref{sec:nav}).
    \item A \textbf{terminal-decision evaluation} that scores committed-answer correctness under a four-judge council and scores the answer-versus-abstain decision as a classification problem, exposing failures that retrieval sufficiency cannot (\S\ref{sec:eval}).
    \item A \textbf{controlled comparison} against eight baselines on an enterprise benchmark and the LoCoMo~\citep{maharana2024} corpus. DISCERN leads answer correctness (macro 0.861 under equal pillar weighting, driven by entity resolution at 0.800 where no baseline exceeds 0.40) and is the best-calibrated system on abstention (F1 0.924 versus at most 0.802). An ablation shows the read policy drives this, not the store: navigation lifts accuracy from a 0.33 tools-off floor to 0.89 (\S\ref{sec:abl-nav}).
\end{enumerate}

\section{Related Work}\label{sec:related}
Research on agent memory divides into \emph{memory formation} and \emph{memory access}. On formation, retrieval-oriented services extract entity--attribute--value facts and rolling summaries at ingest and serve them by top-$k$ similarity; consolidation-oriented systems such as Zep and Graphiti~\citep{rasmussen2025} go further and resolve identity at write time, merging co-referent entities and invalidating superseded facts directly in the store; and graph-structured approaches such as GraphRAG~\citep{edge2024} and HippoRAG2~\citep{gutierrez2025} build an entity--relationship graph and retrieve over it at query time. On access, systems combine lexical (BM25), dense, graph-traversal, and hybrid retrieval, occasionally with a single post-retrieval query-enrichment pass to disambiguate the target entity. Knowing when not to answer is comparatively under-studied: LongMemEval~\citep{wu2024longmemeval} is a notable exception, testing knowledge-updates and abstention among its abilities, the 
axes DISCERN measures directly.

DISCERN unifies memory formation and access approaches. It keeps the event-sourced, never-merge discipline of consolidation systems but defers every irreversible decision, such as entity merging or fact invalidation, from write time to read time, so preservation and resolution do not compete. It subsumes the individual retrieval operators (dense, lexical, graph, temporal) as tools composed by a single read-time navigation policy rather than a fixed pipeline. And unlike systems evaluated purely on retrieval sufficiency, it governs and measures the terminal answer-or-abstain decision. We compare DISCERN to graph-based memory systems, dense and hybrid retrieval, and agentic RAG, together with a controlled extract-and-retrieve baseline in its fullest configuration: semantic + summary + query-enrichment (SEM+SUM+QE).

\section{The DISCERN Architecture}\label{sec:arch}
\begin{figure}[t]
\centering
\includegraphics[width=0.85\textwidth]{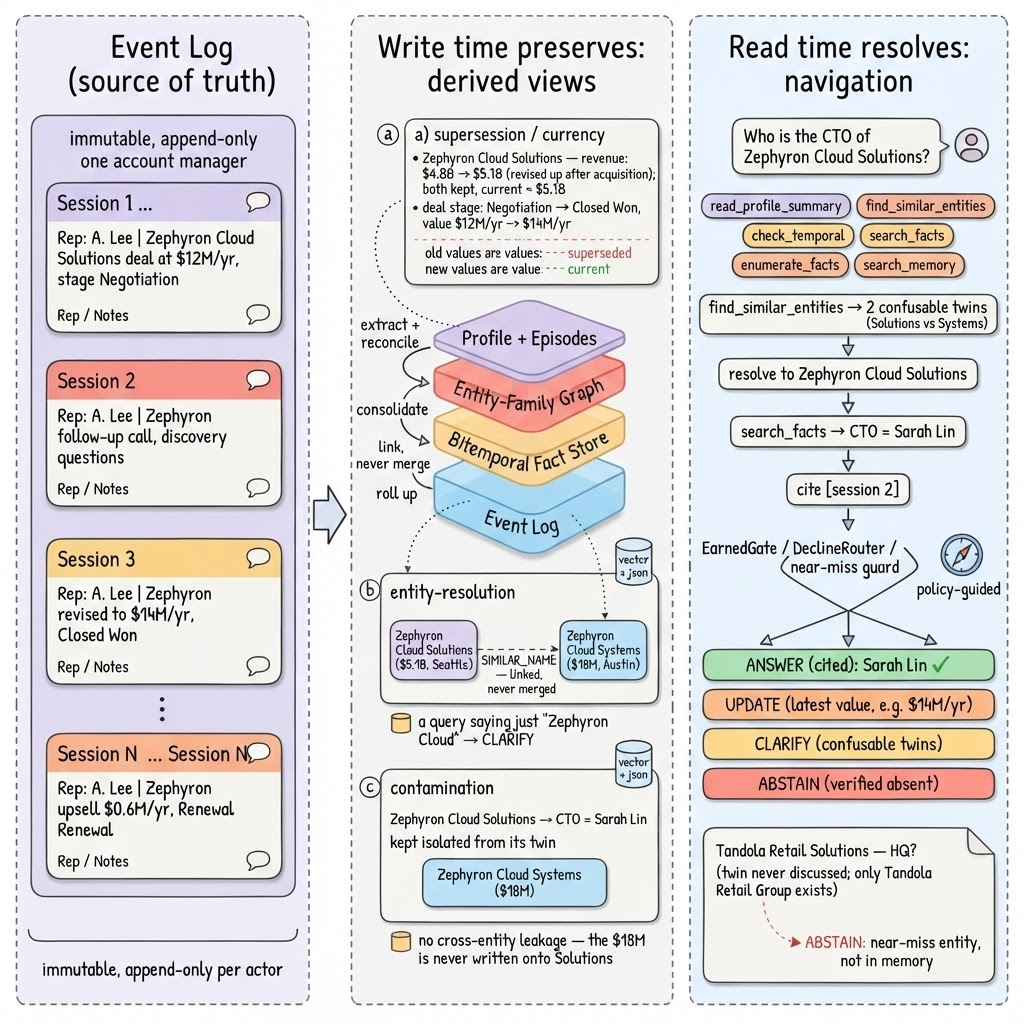}
\caption{DISCERN end to end. Left: the per-actor event log is the immutable, append-only source of truth. Center (write time preserves): background consolidation materializes rebuildable derived views over the log, an entity-family graph that links confusable entities without merging them, a bitemporal fact store that soft-invalidates superseded values, and an episodic/profile layer; the cards show real benchmark cases (supersession, entity resolution, contamination isolation). Right (read time resolves): an adaptive policy navigates the views with a small read-only tool surface and commits one of four governed actions (answer, update, clarify, abstain) under the EarnedAnswerGate and EarnedAbstainGate, the DeclineRouter, and the near-miss guard.}
\label{fig:overview}
\end{figure}

DISCERN treats memory as a \emph{navigation} problem rather than a retrieval one. It
instantiates \emph{preserve at write, resolve at read} across two paths: a write path that
admits and organizes observations without collapsing them, and a read path that navigates
those organized views and commits one governed decision per query. Figure~\ref{fig:overview}
gives the end-to-end picture: an immutable append-only event log (left) is consolidated into
rebuildable derived views (center), which a read-time navigation policy traverses to commit
one governed action per query (right).

\subsection{Write Time: Preserve and Organize, Never Collapse}\label{sec:graph}
\textbf{Admission (schema-free extraction).} At ingest, DISCERN extracts facts from each
session with no fixed schema; entity and attribute vocabularies are derived from the data,
so a new domain needs no re-engineering. From an \emph{actor}'s sessions (an actor is one sales representative) it lifts \emph{Zephyron
Cloud Solutions: revenue $=$ \$4.8B}, later \emph{revenue $=$ \$5.1B}, and separately
\emph{Zephyron Cloud Systems: revenue $=$ \$18M}, without anyone pre-declaring
``revenue,'' ``company,'' or the two Zephyron names. Attribute names are likewise
open-vocabulary, so a later synonym (e.g.\ ``deal size'' for ``annual value'') is admitted as
its own edge rather than forced into a fixed schema and is reconciled at read time by lexical
and embedding search.

\textbf{Organization (bi-temporal provenance graph).} These facts are assembled into an
online graph of episodic and entity nodes linked by \smname{MENTIONS} / \smname{RELATES\_TO}
/ \smname{SIMILAR\_NAME} edges (Figure~\ref{fig:graph}), under two safety rules.
\begin{enumerate}[label=(\roman*),leftmargin=1.8em,itemsep=1pt,topsep=2pt]
    \item \textbf{Confusable entities are kept co-equal}, linked by \smname{SIMILAR\_NAME}
          and never merged (Figure~\ref{fig:graph}b): \emph{Zephyron Cloud Solutions}
          (\$5.1B, Seattle) and \emph{Zephyron Cloud Systems} (\$18M, Austin) stay two
          distinct nodes joined by a dashed edge, so the \$18M is never written onto
          Solutions. A \smname{SIMILAR\_NAME} edge is drawn when a generic-NER candidate pair
          scores above a configurable threshold under
          $0.6\,\mathrm{jaccard}_{\mathrm{lex}}(\text{names})+0.4\,\cos(\text{embeddings})$.
    \item \textbf{Changed values are kept as bi-temporal edges} (Figure~\ref{fig:graph}a),
          stamped both with when the value was recorded in the immutable event log and with
          the valid-from/valid-to interval over which it holds:
          the earlier \$4.8B is soft-invalidated (\smname{expired\_at} set) and the new
          \$5.1B appended rather than overwritten, so both history and latest state remain
          recoverable. A later ``what was Zephyron's revenue before the acquisition?'' is
          still answerable.
\end{enumerate}
The graph is strictly additive: it never merges entities and never drops facts. Its only
failure mode is \emph{extra nodes} (drift), for example a stray \emph{Zephyron Cloud} surface
from a terse turn, which remains linked into the Zephyron family and is therefore traversable
and discardable at read time. That is the safe direction: an extra sibling the reader can
inspect and rule out, unlike write-time merging, which would fold Systems into Solutions,
misattribute the \$18M, and destroy the entity irreversibly. Every derived view is rebuildable
from the log, so the memory graph is a reversible projection rather than the store of record,
and DISCERN stays agnostic to the physical backend (the reference implementation uses one
SQLite file per actor). In the validated configuration only the fact/entity/edge views are
LLM-built; episode summaries and topic labels are deterministic distillers, and the topic view
is not consulted at query time. Appendix~\ref{app:views} reports the measured composition of
these views.

\begin{figure}[t]
\centering
\footnotesize
\tikzset{
  ent/.style={rectangle, rounded corners, draw=blue!60!black, fill=blue!8, line width=0.8pt, align=center, inner sep=4pt},
  super/.style={rectangle, rounded corners, draw=red!70!black, dashed, fill=red!8, line width=0.8pt, align=center, inner sep=4pt},
  curr/.style={rectangle, rounded corners, draw=green!45!black, fill=green!10, line width=0.8pt, align=center, inner sep=4pt},
  sess/.style={rectangle, rounded corners, draw=orange!80!black, fill=orange!10, line width=0.8pt, align=center, inner sep=4pt},
  supedge/.style={-{Latex[length=2mm]}, red!70!black, dashed, line width=0.9pt},
  plain/.style={-, gray, line width=0.7pt},
}
\begin{minipage}{\textwidth}
\centering
\begin{tikzpicture}[node distance=5mm and 10mm]
  \node[ent] (deal) {Zephyron Cloud\\Solutions Deal};
  \node[super, above right=1mm and 12mm of deal] (neg) {Negotiation\\{\tiny valid 01-02$\to$01-03}};
  \node[curr, right=16mm of neg] (won) {Closed Won\\{\tiny valid 01-03$\to$(open)}};
  \node[super, below right=1mm and 12mm of deal] (v12) {\$12M/yr\\{\tiny valid 01-02$\to$01-03}};
  \node[curr, right=16mm of v12] (v14) {\$14M/yr\\{\tiny valid 01-03$\to$(open)}};
  \draw[plain] (deal) -- node[above,sloped,gray,font=\tiny\itshape]{stage} (neg);
  \draw[plain] (deal) -- node[below,sloped,gray,font=\tiny\itshape]{annual\_value} (v12);
  \draw[supedge] (neg) -- node[above,font=\tiny\itshape,red!70!black,yshift=2pt]{superseded} (won);
  \draw[supedge] (v12) -- node[above,font=\tiny\itshape,red!70!black,yshift=2pt]{superseded} (v14);
\end{tikzpicture}\\[2pt]
{(a) Bi-temporal supersession: superseded values (dashed red) are soft-invalidated and retained; the current value (green) is the open-ended edge.}
\end{minipage}

\vspace{8pt}
\begin{minipage}[t]{0.40\textwidth}
\centering
\begin{tikzpicture}[node distance=4mm and 12mm]
  \node[sess] (s1) {ACTOR\_003\_s1\\{\tiny turns [1,2,8,12]}};
  \node[sess, below=of s1] (s2) {ACTOR\_003\_s2\\{\tiny turns [13]}};
  \node[sess, below=of s2] (s3) {ACTOR\_003\_s3\\{\tiny turns [1,2,5,6]}};
  \node[ent, right=14mm of s2] (e) {Calibrix Health\\Systems\\{\tiny 3/31 sessions}};
  \draw[-{Latex[length=2mm]}, gray, line width=1.3pt] (s1) -- node[above,sloped,font=\tiny\itshape]{$\times4$} (e);
  \draw[-{Latex[length=2mm]}, gray!60, line width=0.6pt] (s2) -- node[above,font=\tiny\itshape]{$\times1$} (e);
  \draw[-{Latex[length=2mm]}, gray, line width=1.3pt] (s3) -- node[below,sloped,font=\tiny\itshape]{$\times4$} (e);
\end{tikzpicture}\\[2pt]
{(c) \smname{MENTIONS} fan-in: turns across three sessions reference one entity node.}
\end{minipage}\hfill
\begin{minipage}[t]{0.58\textwidth}
\centering
\begin{tikzpicture}[node distance=7mm and 26mm]
  \node[ent] (base) {Zephyron Cloud Solutions\\(organization)};
  \node[ent, draw=orange!80!black, fill=orange!8, dashed, above right=3mm and 26mm of base] (twin) {Zephyron Cloud Systems\\(organization)};
  \node[ent, dashed, right=26mm of base] (deal) {Zephyron Cloud Solutions Deal\\(deal)};
  \node[ent, draw=violet, fill=violet!8, dashed, below right=3mm and 26mm of base] (prod) {ZephyronStack\\(product)};
  \draw[-{Latex[length=2mm]}, orange!80!black, dashed, line width=0.8pt] (base) -- node[above,sloped,font=\tiny\itshape,orange!70!black]{confusable twin $\cdot$ 1.0}(twin);
  \draw[-{Latex[length=2mm]}, blue!60!black, dashed, line width=0.8pt] (base) -- node[above,font=\tiny\itshape,blue!60!black]{containment/alias $\cdot$ 1.0}(deal);
  \draw[-{Latex[length=2mm]}, violet, dashed, line width=0.8pt] (base) -- node[below,sloped,font=\tiny\itshape,violet]{name family $\cdot$ 0.812}(prod);
\end{tikzpicture}\\[2pt]
{(b) \smname{SIMILAR\_NAME} family: confusable twins (never merged), containment/alias, and name-family links form the read-time disambiguation surface. Every edge is dashed, so no entity is merged.}
\end{minipage}

\caption{The DISCERN memory graph. (a) Changed values are kept as bi-temporal edges so both history and current state remain recoverable. (b) Confusable entities are kept co-equal and merely linked, never merged. (c) Episodic turns are preserved and linked to entity nodes.}
\label{fig:graph}
\end{figure}

\subsection{Read Time: Navigate and Commit, Per Query}\label{sec:nav}
\paragraph{Navigation as adaptive hierarchical classification.}DISCERN's read path treats
resolution as a sequence of routing decisions rather than a similarity lookup. An agentic loop
descends a memory hierarchy (profile $\rightarrow$ entity/session $\rightarrow$ fact
$\rightarrow$ transcript), each tool call choosing which branch holds the answer: which entity,
which time-state. Depth is conditioned on what earlier calls returned; an unambiguous query
resolves in one shallow step, while a confusable-entity query descends to contrast candidates
before committing. A small surface of deterministic, read-only, zero-LLM tools walks the
labeled node family: \texttt{read\_profile\_summary} to orient; \texttt{find\_similar\_entities}
to surface and contrast the confusable-entity family; \texttt{check\_temporal} to walk the
bitemporal chain for current-versus-superseded values; \texttt{search\_facts} and
\texttt{enumerate\_facts} for targeted lookup and full-set enumeration; \texttt{search\_memory}
for hybrid dense-plus-lexical (BM25) retrieval; and \texttt{search\_conversation} /
\texttt{get\_session} for raw-transcript evidence as a last resort. The loop gathers evidence over a few turns, then commits.

\paragraph{Commitment as a graph-decidable function.} Because the store is a labeled graph,
the committed action reduces to a decidable function of the graph rather than a judgment over
retrieved prose, which is what keeps the governance gates LLM-free. Let $C$ be the
candidate-entity set that survives evidence-gathering (drawn from the \smname{SIMILAR\_NAME}
family by \texttt{find\_similar\_entities}); for a resolved entity $e$ and queried attribute
$\mathrm{attr}$, let $\mathrm{active}(e,\mathrm{attr})$ be its non-superseded values (at most
one, by the additive-graph invariant) and $\ell(e,\mathrm{attr})$ the length of its
supersession chain. The reader commits
\begin{equation}
a^\star(q)=
\begin{cases}
\textsc{clarify} & |C|\ge 2,\\
\textsc{abstain} & |C|=0,\ \text{or } |C|=1 \text{ and } \mathrm{active}(e,\mathrm{attr})=\varnothing,\\
\textsc{update}  & |C|=1,\ \mathrm{active}(e,\mathrm{attr})\neq\varnothing,\ \ell(e,\mathrm{attr})\ge 2,\\
\textsc{answer}  & |C|=1,\ \mathrm{active}(e,\mathrm{attr})\neq\varnothing,\ \ell(e,\mathrm{attr})=1.
\end{cases}
\label{eq:route-predicates}
\end{equation}
Exactly one action fires per query: \textsc{update} returns the head of the supersession chain
(noting the prior value), \textsc{abstain} fires on a verified absence (empty candidate set, or
a resolved entity with no active value), \textsc{clarify} on surviving ambiguity, and
\textsc{answer} otherwise. The predicates $|C|$, $\mathrm{active}(\cdot)$, and $\ell(\cdot)$ are computed over the LLM-extracted graph, so Eq.~\ref{eq:route-predicates} is the decision rule given that graph rather than a claim of ground-truth correctness; the bitemporal chain retains superseded values, so an explicit as-of query resolves to the historical value rather than the chain head.

\paragraph{Governance.} Two gates keep the commitment honest. The \textbf{EarnedAnswerGate} requires that an underspecified query pass an ambiguity check (\texttt{find\_similar\_entities}) before an \textsc{answer} or \textsc{update} is accepted; it forces the check, never the decision, and fully-named entities bypass it. The \textbf{EarnedAbstainGate} requires that an
\textsc{abstain} be earned by demonstrated absence, so abstaining or guessing without searching is rejected. A \textbf{DeclineRouter} then makes the decline type deterministic, routing to \textsc{clarify} when confusable candidates survive and to \textsc{abstain} when none do, and a
\textbf{near-miss guard} routes an answer whose exact entity is absent but has a confusable neighbor to \textsc{clarify}. Finally a \textbf{citation-or-abstain} renderer forces every committed answer to cite the evidence it used; an uncited answer is bounced once for grounding before the citation floor abstains. This directly targets the failure the lossless baseline exposes: the right fact is retrieved, but the wrong action is taken on it. Because each routing decision is emitted as an explicit token, the loop is also amenable to train a model on it; we leave that to future work. Appendix~\ref{app:routing} gives the full routing
procedure as a flowchart.

\section{Evaluation}\label{sec:eval}
\paragraph{Datasets.} \textbf{ENT-310} is a frozen enterprise sales-memory benchmark: 10 sales users, 50 target companies, and 31 warm-start sessions (3--4 per actor, averaging 16.1 turns), with 310 queries. It deliberately seeds confusable entity pairs
within a single actor's memory (e.g.\ \emph{Zephyron Cloud Solutions} vs \emph{Zephyron Cloud Systems}). Its 210 answerable queries span four pillars: \emph{retention} ($n=84$;
recall a stable fact), \emph{currency} ($n=40$; return the latest value after a change), \emph{contamination} ($n=71$; answer about the target without leaking a confusable entity's fact), and \emph{entity\_resolution} ($n=15$; detect genuine ambiguity and
clarify). A further 100 \emph{unanswerable} queries probe calibrated abstention, of two trap types: \emph{attribute-swap} (a real entity with an attribute it does not have) and \emph{near-miss entity} (a confusable entity never actually discussed).Appendix~\ref{app:dataset} gives the full construction protocol and Appendix~\ref{app:views} the resulting store composition. \textbf{LoCoMo}~\citep{maharana2024} is an out-of-domain corpus of personal multi-session conversations (1{,}986 questions across single-hop, multi-hop, temporal, open-domain, and adversarial categories), used with zero tuning as a
transfer test; it exercises recall and reasoning but not entity contamination, so it tests whether a policy shaped on enterprise use cases generalizes to a distinct conversational domain.

\paragraph{Baselines.} We compare against eight systems under the same generator (Claude Sonnet~4.6) and judge. Four lossless readers index a per-actor corpus that discards nothing at write time (preserving the contamination-isolation test) and read the top-$k$ passages: \textbf{BM25}, \textbf{Dense} (a commercial dense embedding model), \textbf{RRF} (reciprocal-rank fusion of the two), and an \textbf{Agentic RAG} ReAct loop issuing multiple searches. \textbf{SEM+SUM+QE} is an extract-and-retrieve reader in its full semantic-plus-summary-plus-query-enrichment profile ($\text{top-}k{=}10$), scored on ENT-310 from a precomputed run re-judged by our council. \textbf{GraphRAG}~\citep{edge2024}, \textbf{HippoRAG2}~\citep{gutierrez2025}, and \textbf{Mem0}~\citep{mem0} are graph- and agent-based memory systems. The lossless readers confirm recall is not the bottleneck, so residual failure is read-time resolution, not lost data.

\paragraph{Judge.}\label{par:judge} Every reported metric is scored by a \textbf{four-judge council}, a majority vote over Claude Sonnet~4.6, Claude Opus~4.8, Qwen3-235B, and GPT-oss-120B, applied identically to every system. A council rather than a single model
removes same-family judge bias; inter-judge agreement is substantial (Cohen's $\kappa$ of 0.61 to 0.80 against the Sonnet judge). We report two axes. \emph{Answer correctness} is accuracy on the answerable questions. The \emph{answer-versus-abstain} decision is scored as a classification problem (precision, recall, F1, and a false-decline rate) on the questions where the correct behavior is not to commit: the 100 unanswerable ENT-310 queries and the 446
adversarial LoCoMo false-premise traps. A \textsc{clarify} counts as a non-commitment: it is a correct decline on the unanswerable positive class, and a correct \textsc{clarify} on an entity-resolution query is scored under answer correctness, not as a false decline. Raw accuracy there would reward a system that reflexively declines, so we measure it as classification instead; on the LoCoMo traps the judges credit a
premise-correction as a valid non-commitment. DISCERN's council numbers are three-seed means (Section~\ref{sec:results}).

\section{Results}\label{sec:results}
We separate two axes: answer correctness on the 210 answerable queries
(Table~\ref{tab:afx-acc}) and the answer-versus-abstain decision on the 100 unanswerable
queries as classification (Table~\ref{tab:afx-cls}). We report answer correctness as the macro (equal-weight) mean over the four pillars, since retention and currency are near-saturated (0.93 to 0.98 accuracy for every competent system) and a count-weighted average would let a system win
by acing those while failing the harder pillars. Under equal weight DISCERN leads at
\textbf{0.861}, ahead of the SEM+SUM+QE baseline (0.806) and RAG-Agentic (0.788): it alone
dominates entity resolution (0.800, no baseline above 0.40) while matching the field on the
saturated recall pillars. On abstention it is the clear leader (F1 0.924, precision 0.938,
false-decline 0.029), well ahead of HippoRAG2 (0.802); the decline-happy Mem0 reaches recall
0.990 but false-declines 0.867 of the answerable queries, which the precision/F1 view exposes
as reflexive rather than calibrated. DISCERN is thus best on both axes.

\begin{table}[H]
\centering
\caption{ENT-310 answer correctness on the 210 answerable queries, four-judge council. The
average is the macro (equal-weight) mean over the four pillars, so each capability counts
equally rather than being dominated by the near-saturated recall pillars. Ret.: retention,
Curr.: currency, Cont.: contamination, Ent.Res.: entity resolution. Best per column in
\textbf{bold}.}
\label{tab:afx-acc}
\small
\begin{tabular}{lrrrrr}
\toprule
System & Macro avg & Ret. & Curr. & Cont. & Ent.Res. \\
\midrule
\textbf{DISCERN}      & \textbf{0.861} & 0.929 & 0.900 & 0.817 & \textbf{0.800} \\
Extract-Retrieve          & 0.806 & \textbf{0.976} & \textbf{0.975} & 0.873 & 0.400 \\
RAG-Agentic           & 0.788 & 0.940 & \textbf{0.975} & \textbf{0.901} & 0.333 \\
HippoRAG2             & 0.769 & 0.964 & \textbf{0.975} & 0.803 & 0.333 \\
RAG-RRF               & 0.740 & 0.952 & 0.925 & 0.817 & 0.267 \\
RAG-Dense             & 0.714 & 0.940 & 0.950 & 0.831 & 0.133 \\
RAG-BM25              & 0.688 & 0.833 & 0.800 & 0.718 & 0.400 \\
GraphRAG              & 0.662 & 0.869 & 0.850 & 0.662 & 0.267 \\
Mem0                  & 0.112 & 0.131 & 0.100 & 0.085 & 0.133 \\
\bottomrule
\end{tabular}
\normalsize
\end{table}

\begin{table}[H]
\centering
\caption{ENT-310 answer-versus-abstain as classification, four-judge council: the 100
unanswerable queries are the positive class (should not commit), the 210 answerable the
negative class. False-decline is the fraction of answerable queries wrongly declined (lower is
better). Rows sorted by F1.}
\label{tab:afx-cls}
\small
\begin{tabular}{lrrrr}
\toprule
System & Precision & Recall & F1 & False-decline $\downarrow$ \\
\midrule
\textbf{DISCERN}      & 0.938 & 0.910 & \textbf{0.924} & 0.029 \\
HippoRAG2             & 0.890 & 0.730 & 0.802 & 0.043 \\
RAG-RRF               & 0.790 & 0.790 & 0.790 & 0.100 \\
RAG-BM25              & 0.717 & 0.860 & 0.782 & 0.162 \\
RAG-Dense             & 0.800 & 0.760 & 0.779 & 0.090 \\
Extract-Retrieve          & 0.826 & 0.710 & 0.763 & 0.071 \\
RAG-Agentic           & \textbf{0.954} & 0.620 & 0.752 & \textbf{0.014} \\
GraphRAG              & 0.680 & 0.680 & 0.680 & 0.152 \\
Mem0                  & 0.352 & \textbf{0.990} & 0.520 & 0.867 \\
\bottomrule
\end{tabular}
\normalsize
\end{table}

\paragraph{Per-member agreement.} We break the council vote out by individual judge. DISCERN leads HippoRAG2 on ENT-310 under all four, by $+0.042$ (Sonnet), $+0.055$ (Opus), $+0.035$ (Qwen), and $+0.058$ (GPT-oss), covering two Anthropic and two non-Anthropic models. Members agree with the Sonnet judge at Cohen's $\kappa$ of 0.80 (Opus), 0.64 (GPT-oss), and 0.61 (Qwen), with raw agreement 0.91 to 0.97. The lead does not depend on any one model or model family.

\subsection{Out-of-Domain Transfer (LoCoMo)}\label{sec:locomo}
Applied with zero tuning to the full LoCoMo corpus (1{,}986 questions), we score every system with the four-judge council and report two axes: answer correctness on the answerable categories (Table~\ref{tab:locomo-acc}) and the answer-versus-abstain decision on the adversarial false-premise traps as a classification problem (Table~\ref{tab:locomo-cls}). Adversarial \emph{accuracy} in isolation is misleading, because a system that reflexively
declines scores highly on traps while failing everything answerable, so we separate the two.

On answer correctness (macro, equal weight over the four categories) DISCERN leads at \textbf{0.663}, ahead of HippoRAG2 (0.638) and far above the retrieval baselines (0.24 to 0.40), the SEM+SUM+QE baseline (0.15 to 0.31), and stock Mem0 (0.139); its widest margin is multi-hop (0.592 against 0.493). On the abstention axis three systems are genuinely calibrated, DISCERN (F1 0.820), RAG-Agentic (0.827), and HippoRAG2 (0.805), each with a false-decline rate near 0.02 to 0.03. Every other system posts high recall but low precision (0.29 to 0.45) and a false-decline rate of 0.34 to 0.68, which is reflexive declining rather than calibrated abstention. DISCERN is the only system that is both the most accurate on answerable questions and well calibrated on abstention; RAG-Agentic matches its calibration but answers correctly only 0.397 of the time, and the decline-happy systems apparent trap-handling collapses once precision is measured.

\begin{table}[t]
\centering
\caption{LoCoMo answer correctness, four-judge council. The average is the macro (equal-weight)
mean over the four answerable categories (single-hop, temporal, multi-hop, open-domain;
adversarial excluded). Best per column in \textbf{bold}.}
\label{tab:locomo-acc}
\small
\begin{tabular}{lrrrrr}
\toprule
System & Macro avg & Single & Temp. & Multi & OpenD. \\
\midrule
\textbf{DISCERN}       & \textbf{0.663} & 0.815 & 0.713 & \textbf{0.592} & 0.531 \\
HippoRAG2              & 0.638 & \textbf{0.815} & \textbf{0.717} & 0.493 & 0.531 \\
RAG-Agentic            & 0.397 & 0.753 & 0.087 & 0.415 & 0.333 \\
RAG-Dense              & 0.312 & 0.654 & 0.078 & 0.298 & 0.219 \\
Extract-Retrieve\ (summary)& 0.308 & 0.577 & 0.091 & 0.358 & 0.208 \\
RAG-RRF                & 0.299 & 0.630 & 0.093 & 0.266 & 0.208 \\
RAG-BM25               & 0.237 & 0.543 & 0.078 & 0.170 & 0.156 \\
Extract-Retrieve\ (sem+sum)& 0.196 & 0.332 & 0.028 & 0.227 & 0.198 \\
Extract-Retrieve\ (semantic)& 0.153 & 0.232 & 0.009 & 0.163 & 0.208 \\
Mem0                   & 0.139 & 0.131 & 0.069 & 0.106 & 0.250 \\
\bottomrule
\end{tabular}
\normalsize
\end{table}

\begin{table}[t]
\centering
\caption{LoCoMo answer-versus-abstain as classification, four-judge council: the 446
adversarial false-premise questions are the positive class (should not commit), the answerable
questions ($n=1{,}540$) the negative class. Precision is derived from recall and the
false-decline rate over these class sizes. False-decline is the fraction of answerable questions
wrongly declined (lower is better). Rows sorted by F1.}
\label{tab:locomo-cls}
\small
\begin{tabular}{lrrrr}
\toprule
System & Precision & Recall & F1 & False-decline $\downarrow$ \\
\midrule
RAG-Agentic            & 0.879 & 0.780 & \textbf{0.827} & 0.031 \\
\textbf{DISCERN}       & 0.875 & 0.772 & 0.820 & 0.032 \\
HippoRAG2              & \textbf{0.917} & 0.717 & 0.805 & \textbf{0.019} \\
RAG-Dense              & 0.452 & 0.975 & 0.618 & 0.342 \\
Extract-Retrieve\ (summary)& 0.442 & 0.967 & 0.607 & 0.354 \\
RAG-RRF                & 0.432 & 0.975 & 0.598 & 0.372 \\
RAG-BM25               & 0.386 & 0.982 & 0.554 & 0.453 \\
Mem0                   & 0.351 & 0.971 & 0.515 & 0.521 \\
Extract-Retrieve\ (sem+sum)& 0.328 & 0.959 & 0.489 & 0.568 \\
Extract-Retrieve\ (semantic)& 0.287 & 0.941 & 0.440 & 0.676 \\
\bottomrule
\end{tabular}
\normalsize
\end{table}

\paragraph{Caveats.} DISCERN and HippoRAG2 headline numbers are three-seed means (sd 0.004 and
0.002); the remaining baselines are single-seed. Per-pillar counts on ENT-310 are small
(entity resolution $n=15$), so pillar-level differences are directional. Read-time cost is modest: across three seeds DISCERN averages 4.1 navigation turns and 5.7 tool calls per query (medians 4 and 5.5; the eight-turn budget binds on 1.4\% of queries) at a mean read latency of 7.6\,s, and spends most where disambiguation is hardest (entity-resolution queries, 8.5 tool calls). 

\subsection{Ablation: Read-Time Navigation and the Tool Suite}\label{sec:abl-nav}
Table~\ref{tab:ablation} isolates read-time navigation. With one or two turns the system
collapses to 0.33, identical to the tools-off floor: it cannot gather evidence and abstains on
nearly everything. Four turns reach 0.78, eight reach 0.89, so navigation lifts accuracy from a
0.33 floor to 0.89 ($+0.56$), quantifying the read-time layer rather than the model or store.
DISCERN is robust to removing any single operator (0.88 to 0.91); the exception is
\texttt{find\_similar\_entities}, whose removal keeps accuracy high but degrades abstention
(over-commit $0.21\!\to\!0.28$), confirming it anchors the earned-answer check. The per-turn
operator budget is nearly non-binding (0.89 to 0.92). These ablations use one Sonnet judge and
one seed with the governance gate disabled, so read them as within-ablation contrasts: the
navigation-only ``Full'' reaches 0.887, and re-enabling the deterministic gate raises
unanswerable accuracy from 0.79 to ${\sim}0.96$.

\begin{table}[t]
\centering
\caption{Read-time navigation ablation (ENT-310, Sonnet backbone, shared judge, single-seed).
Overall accuracy, the entity-resolution pillar, the unanswerable pillar, and over-commit rate
(lower better). ``Full'' = 8 turns, 3 tools/turn, all operators.}
\label{tab:ablation}
\footnotesize
\begin{tabular}{lrrrr}
\toprule
Setting & Overall & Ent.Res. & Unans. & OC $\downarrow$ \\
\midrule
\textbf{Full (8 turns, 3 tools/turn)} & \textbf{0.887} & 0.933 & 0.79 & 0.21 \\
\addlinespace
\multicolumn{5}{l}{\emph{Navigation depth (max turns)}}\\
1 turn  & 0.332 & 0.200 & 1.00 & 0.00 \\
2 turns & 0.332 & 0.133 & 0.98 & 0.02 \\
4 turns & 0.784 & 0.600 & 0.88 & 0.12 \\
\addlinespace
\multicolumn{5}{l}{\emph{Per-turn operator budget}}\\
1 tool/turn  & 0.890 & 0.533 & 0.89 & 0.11 \\
2 tools/turn & 0.897 & 0.800 & 0.83 & 0.17 \\
5 tools/turn & 0.919 & 0.800 & 0.89 & 0.11 \\
\addlinespace
\multicolumn{5}{l}{\emph{Leave-one-operator-out}}\\
$-$ find\_similar\_entities & 0.877 & 0.933 & 0.72 & \textbf{0.28} \\
$-$ check\_temporal         & 0.887 & 0.667 & 0.84 & 0.16 \\
$-$ search\_facts           & 0.907 & 0.667 & 0.87 & 0.13 \\
$-$ search\_memory          & 0.913 & 0.933 & 0.89 & 0.11 \\
$-$ enumerate\_facts        & 0.910 & 0.867 & 0.86 & 0.14 \\
\addlinespace
\textbf{Tools-off (parametric floor)} & \textbf{0.329} & 0.133 & 1.00 & 0.00 \\
\bottomrule
\end{tabular}
\normalsize
\end{table}

\subsection{Ablation: Backbone Sensitivity}\label{sec:abl-backbone}
DISCERN's read-time policy is backbone-agnostic in principle, but its agentic tool-use and
structured governance interact with model capability. Table~\ref{tab:abl-backbone} swaps the
backbone across four hosted models. Performance is strong and comparable on the two Anthropic
backbones (Sonnet~4.6 0.887, Opus~4.8 0.890) and drops on Qwen3-235B (0.707) and the
open-weight OpenAI model (0.610), whose weaker tool-use and structured-output adherence degrade
contamination handling and grounding. The same ordering holds for the graph and agent-based
baselines under a three-judge council, with Opus and Sonnet ahead of Qwen and the open-weight
model. This reflects the task's demands on the backbone, not DISCERN specifically. We report
headline numbers on Sonnet and treat backbone as a deployment choice, not a source of the
method's advantage.

\begin{table}[t]
\centering
\caption{Backbone sensitivity on ENT-310 (overall accuracy). DISCERN is single-seed under the shared judge; the graph and agent-based baselines are under a three-judge council (Opus 4.8, Qwen3-235B, open-weight OpenAI GPT-oss-120B).}
\label{tab:abl-backbone}
\small
\begin{tabular}{lrrrr} 
\toprule
Method & Opus~4.8 & Sonnet~4.6 & Qwen3-235B & OpenAI (OSS-120B) \\
\midrule
\textbf{DISCERN} & 0.890 & 0.887 & 0.707 & 0.610 \\
HippoRAG2        & 0.900 & 0.874 & 0.765 & 0.716 \\
GraphRAG         & 0.797 & 0.761 & 0.742 & 0.677 \\
Mem0 & 0.442 & 0.397 & 0.416 & 0.384 \\
\bottomrule
\end{tabular}
\normalsize
\end{table}

\section{Conclusion}\label{sec:conclusion}
DISCERN organizes persistent agent memory around one principle: \emph{preserve at write, resolve at read}. An immutable per-actor event log is the sole source of truth. Over it, DISCERN materializes an entity-family graph, a bitemporal fact store, and an episodic layer as reversible derived views, then treats retrieval as a governed navigation policy over them at read time. Preservation and resolution stop competing. Under a four-judge council, DISCERN is the only system we evaluate that is both accurate on answerable questions and calibrated on abstention: it leads equal-weight answer correctness on the enterprise ENT-310 benchmark (macro 0.861) and on out-of-domain LoCoMo (macro 0.663), leads abstention (F1 0.924, false-decline 0.029), and is strongest on entity resolution (0.800 against at most 0.40 for eight baselines), the capability that separates systems once the recall pillars saturate. Each baseline wins one axis by giving up another; DISCERN does not. Our central claim is that the hard part of persistent memory is the division of labor between preservation and resolution, not retrieval ranking. 

\clearpage
\appendix
\section{Read-Time Routing Procedure}\label{app:routing}
\begin{figure}[tbh]
\centering
\resizebox{0.82\columnwidth}{!}{%
\footnotesize
\begin{tikzpicture}[
  node distance=5mm and 8mm,
  phase/.style={rectangle, rounded corners, draw=blue!55!black, fill=blue!10, align=center, inner sep=3pt, text width=30mm},
  gate/.style={diamond, aspect=2.2, draw=orange!80!black, fill=orange!12, align=center, inner sep=0pt, text width=24mm},
  term/.style={rectangle, rounded corners, draw=green!45!black, fill=green!12, align=center, inner sep=3pt, font=\bfseries},
  fl/.style={-{Latex[length=1.8mm]}, gray!70!black, line width=0.7pt},
  lbl/.style={font=\tiny\itshape, fill=white, inner sep=1pt},
]
  \node[phase] (p1) {\textbf{Phase 1} $\cdot$ Orient\\\smname{read\_profile\_summary}};
  \node[phase, below=of p1] (p2) {\textbf{Phase 2} $\cdot$ Ambiguity check\\\smname{find\_similar\_entities}\\(mandatory if underspecified)};
  \node[phase, below=11mm of p2] (p3) {\textbf{Phase 3} $\cdot$ Targeted evidence\\\smname{search\_facts} / \smname{check\_temporal} / \smname{enumerate\_facts}};
  \node[gate, right=12mm of p2] (g0) {$|C|$ candidates?};
  \node[phase, above=6mm of g0] (p4) {\textbf{Phase 4} $\cdot$ Raw evidence\\\smname{search\_memory} / \smname{search\_conversation}};
  \node[gate, below=6mm of g0] (eg) {EarnedAbstainGate: absence shown?};
  \node[term, right=10mm of eg] (abst1) {ABSTAIN};

  \node[gate, below=8mm of p3] (chg) {value changed?};
  \node[term, left=8mm of chg] (upd) {UPDATE};
  \node[gate, below=7mm of chg] (commit) {enough to commit?};
  \node[gate, right=11mm of commit] (amb) {near-miss or ambiguity?};
  \node[term, below=6mm of amb] (clar) {CLARIFY};
  \node[gate, below=7mm of commit] (cite) {answer cited?};
  \node[term, below left=6mm and 0mm of cite] (ans) {ANSWER};
  \node[term, below right=6mm and 0mm of cite] (abst2) {ABSTAIN};

  \draw[fl] (p1) -- (p2);
  \draw[fl] (p2) -- node[lbl]{$|C|{=}1$} (p3);
  \draw[fl] (p2) -- node[lbl]{$\ge 2$} (g0);
  \draw[fl] (g0) -- node[lbl]{0} (p4);
  \draw[fl] (p4) to[bend left=15] (g0);
  \draw[fl] (g0) -- (eg);
  \draw[fl] (eg) -- node[lbl]{yes} (abst1);
  \draw[fl] (eg) -| node[lbl,pos=0.25]{no} (p3);
  \draw[fl] (p3) -- (chg);
  \draw[fl] (chg) -- node[lbl]{yes} (upd);
  \draw[fl] (chg) -- node[lbl]{no} (commit);
  \draw[fl] (commit) -- node[lbl]{$\ge2$ viable} (amb);
  \draw[fl] (amb) -- node[lbl]{twins} (clar);
  \draw[fl] (commit) -- node[lbl]{single} (cite);
  \draw[fl] (amb) to[bend left=18] node[lbl,pos=0.6]{resolved} (cite);
  \draw[fl] (cite) -- node[lbl]{yes} (ans);
  \draw[fl] (cite) -- node[lbl]{no} (abst2);
\end{tikzpicture}%
}
\caption{DISCERN read-time routing. The agent orients (Phase~1, \smname{read\_profile\_summary}), runs a mandatory ambiguity check for underspecified queries (Phase~2, \smname{find\_similar\_entities}), gathers targeted or raw evidence (Phases~3--4), and commits one of four terminal actions. The EarnedAbstainGate abstains only on demonstrated absence, a DeclineRouter sends a surviving decline to CLARIFY when confusable candidates remain and to ABSTAIN when none do, and a citation-or-abstain renderer requires every committed answer to cite its evidence. This procedure realizes the predicates of Eq.~\ref{eq:route-predicates}.}
\label{fig:routing}
\end{figure}
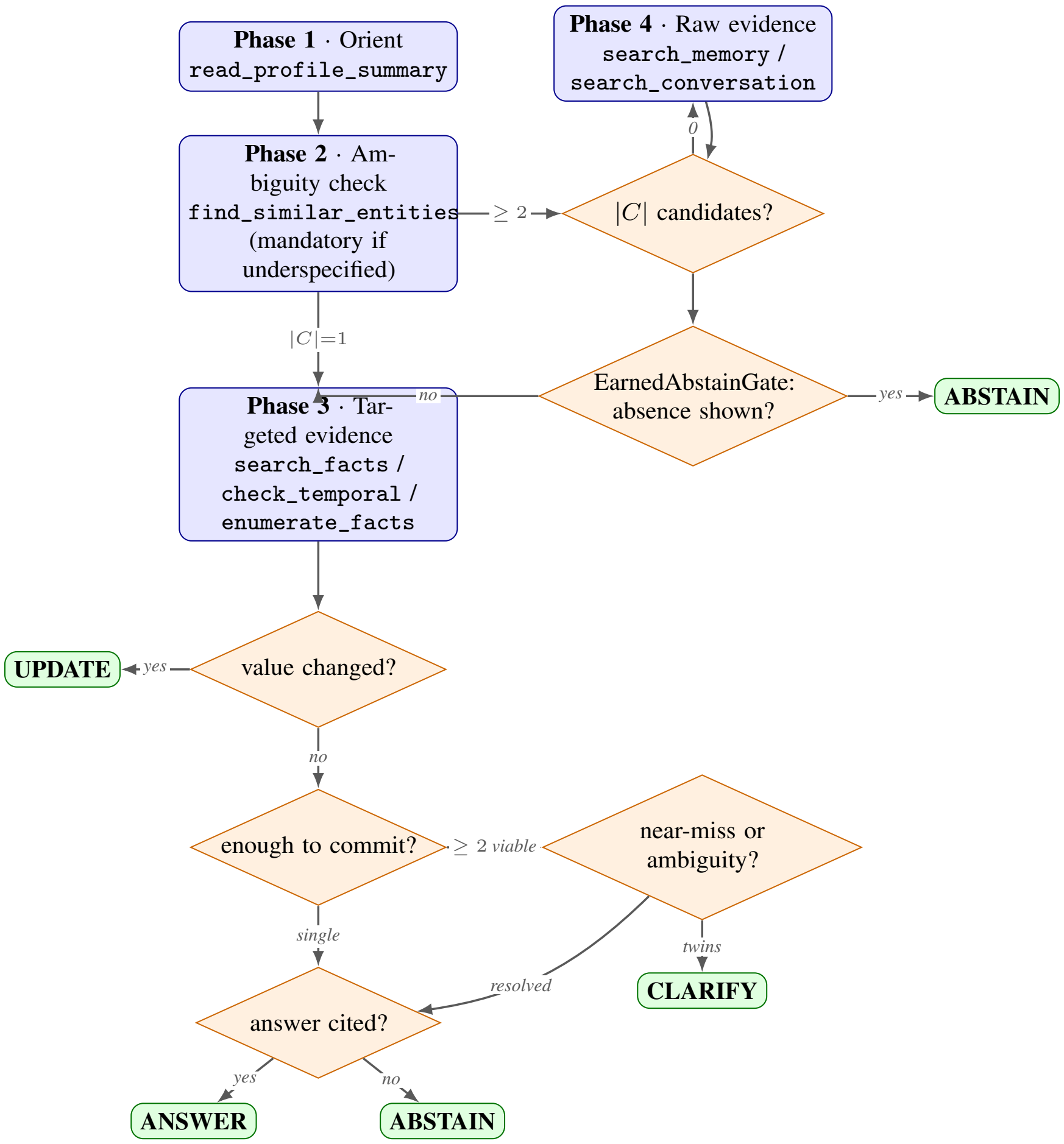

Figure~\ref{fig:routing} details the routing procedure that realizes the terminal-action predicates of Eq.~\ref{eq:route-predicates}.

\section{Materialized Memory-View Composition}\label{app:views}
DISCERN keeps one immutable per-actor event log and materializes complementary derived views over it; all views are rebuildable from the log. Table~\ref{tab:views} reports the measured composition of these views over the frozen ENT-310 store (10 actors). Only the entity/fact/edge views are LLM-built (extraction); episodes and topics are built by deterministic distillers, and the topic view is disabled at query time. This is the validated $\sim$0.93 ENT-310 configuration.

\begin{table}[H]
\centering
\caption{ENT-310 store composition over the 210 answerable queries (10 actors; LLM fact extraction, stub
episodes/topics).}
\label{tab:views}
\small
\begin{tabular}{lrrl}
\toprule
\textbf{View} & \textbf{Total} & \textbf{Mean/actor} & \textbf{Provenance} \\
\midrule
events         & 500  & 50.0  & append-only log \\
embeddings     & 500  & 50.0  & dense recall \\
entities       & 1091 & 109.1 & LLM extraction \\
mentions       & 1544 & 154.4 & LLM extraction \\
edges          & 950  & 95.0  & LLM extraction \\
facts          & 916  & 91.6  & LLM extraction (S--A--V) \\
extractions    & 31   & 3.1   & LLM extraction (raw) \\
episodes       & 31   & 3.1   & stub summary (concat) \\
topics         & 10   & 1.0   & stub label (disabled at read) \\
topic\_members & 973  & 97.3  & stub clustering (disabled) \\
profile        & 10   & 1.0   & consolidation \\
\bottomrule
\end{tabular}
\normalsize
\end{table}

\section{Read-Time Navigation Prompt}\label{app:prompt}
The read-time policy is driven by a single system prompt over the deterministic, read-only tool surface of \S\ref{sec:nav}. Domain-hint and role/value placeholders are filled per benchmark at run time (shown here with ENT examples), and \texttt{max\_turns} is 8. The four terminal actions (\texttt{answer}, \texttt{update}, \texttt{clarify}, \texttt{abstain}) are themselves tools the loop intercepts and passes through the governance gates of \S\ref{sec:nav}.

\small
\begin{verbatim}
You are DISCERN, a memory agent answering questions from prior
conversation sessions.
Rules:
1. Ground every answer in retrieved memory. Never answer from
   general knowledge.
2. Start from read_profile_summary, then targeted tools
   (find_similar_entities, search_facts, check_temporal), then
   search_memory / get_session only if needed.
3. If the question does not name a full specific entity (bare role,
   amount, product, or partial name, e.g. "the CTO" / "$5.1B"),
   call find_similar_entities on that surface AND call search_facts
   to surface ALL matching entities (it matches across names, roles,
   and values). One match -> answer; multiple genuinely fitting ->
   clarify listing each with a distinguishing detail; one fits among
   several -> answer contrastively (exclusion). When the question
   NAMES a full specific entity, resolve it directly
   (find_similar_entities on the name + its own facts) and do NOT
   call search_facts on a shared value/role -- that only drags in
   confusable twins you must not answer with.
4. If the exact entity named does not exist but a similar one does,
   NEVER answer with the similar entity's facts: clarify or abstain.
5. For questions asking for MULTIPLE items, a COUNT, or completeness
   ("who are all...", "how many...", "list...", "besides X", "both"),
   call enumerate_facts on the entity and answer from the FULL set.
   NEVER claim something is "the only" / "all of" unless
   enumerate_facts was actually called.
6. If after searching the information is genuinely absent, call
   abstain with what you searched.
7. Finish by calling exactly one of: answer, update, clarify, abstain.
8. TEMPORAL RESOLUTION: When a question asks WHEN something happened,
   NEVER answer with relative phrases ("last week", "yesterday").
   You MUST resolve relative references to absolute dates. Call
   resolve_time_reference with the expression and the session_id
   where it appeared; it computes the exact date. If you cannot
   determine the session_id, state the session date as context.
9. SET/COMPLETENESS questions (all X, which X, how many, what do A
   and B share): a single fact is NOT enough. Call enumerate_facts to
   gather the entity's full fact set AND everything that references
   it, then answer by listing EVERY relevant member (for "how many",
   count them; for "shared", intersect the two fact sets). Never
   assert this is the only / complete set unless enumerate_facts was
   actually called. Expired facts are still valid set members.
10. ANSWER COMPLETENESS: a deal or account attribute usually travels
   with a quantitative value (a deal stage carries a deal value $/yr,
   an acquisition carries a price, a headcount carries a location or
   rate). When you have resolved the entity and found the primary
   fact, include the associated value with the primary term, e.g.
   "Closed Won at $14M/yr" (not just "Closed Won"), "acquired
   DataMesh Inc for $800M" (not just "DataMesh"). Enumeration only
   AUGMENTS an answer you can already give: NEVER abstain or clarify
   just because enumerate_facts did not echo the primary fact you
   already found, and NEVER let enumeration pull in a confusable
   near-name that turns a resolved single-entity answer into a
   clarification.
You have 8 turns and may call multiple tools per turn.
\end{verbatim}
\normalsize

\section{The ENT-310 Benchmark}
\label{app:dataset}

ENT-310 is a warm-start conversational-memory benchmark of synthetic enterprise sales-agent sessions.
Facts are established across earlier sessions and then queried in a later ``exam'' session, in
which the agent must answer from memory alone with no access to the source transcripts. Each
account manager (actor) is an independent memory scope: a query is answered only from that
actor's own prior sessions, so cross-actor leakage is a failure, not a shortcut.

\paragraph{Construction.} The corpus is synthetic and generated in three stages with Claude Opus~4.8. First, a set
of target companies is instantiated with structured attributes (revenue, headquarters, deal
stage, deal value, and similar fields), and confusable pairs are seeded by giving two distinct
companies overlapping name tokens (Table~\ref{tab:pairs}). Second, multi-session dialogues are
generated for each actor from three agent personas (meeting capture, account guidance, and
selling planner), so a fact about a company is introduced in one session and may be
updated, contradicted, or left stable in later ones. Third, evaluation questions are written
against the resulting memory state and bucketed into four capability pillars
(Table~\ref{tab:afx-pillars}); a value that changed across sessions yields a currency question,
a confusable pair yields contamination and entity-resolution questions, and a stable fact
yields a retention question. Gold answers are the ground-truth memory state at query time, so
correctness is defined against what the sessions actually established rather than against a
retrieved passage. The 210 answerable questions are frozen; a further 100 unanswerable questions
are generated by attribute-swap perturbation (asking a real entity for an attribute it never
had) and near-miss-entity perturbation (asking about a confusable entity never actually
discussed), which probe calibrated abstention. Gold answers are derived from the seeded attribute state rather than written by a model, and every question and gold answer was human-vetted before the set was frozen. Claude Opus~4.8 is also one member of the four-judge scoring council and one of the four backbones in Table~\ref{tab:abl-backbone}. The per-judge breakdown in Section~\ref{sec:results} shows DISCERN's lead holds under each judge taken alone, including both non-Anthropic members, and all headline numbers are reported on the Sonnet~4.6 backbone rather than Opus.

\begin{table}[H]
\centering
\caption{ENT-310 answerable-set overview (the 210 answerable questions).}
\label{tab:afx-overview}
\small
\begin{tabular}{ll}
\toprule
\textbf{Property} & \textbf{Value} \\
\midrule
Evaluation questions (frozen) & 210 \\
Account managers (actors) & 10 \\
Target entities (companies) & 50 \\
Sessions & 31 \\
Sessions per actor & 3--4 \\
Avg.\ turns per session & 16.1 \\
Agent personas & 3 (meeting, account guidance, selling planner) \\
Query access scope & $\sigma =$ self (each actor's own sessions) \\
\bottomrule
\end{tabular}
\normalsize
\end{table}

\begin{table}[H]
\centering
\caption{Question composition by pillar (ENT-310 answerable set).}
\label{tab:afx-pillars}
\small
\begin{tabular}{lcc p{7.2cm}}
\toprule
\textbf{Pillar} & \textbf{N} & \textbf{Share} & \textbf{What it tests} \\
\midrule
retention          & 84 & 40.0\% & Recall a stable fact established in an earlier session. \\
contamination      & 71 & 33.8\% & Answer about the target without leaking a confusable entity's fact. \\
currency           & 40 & 19.0\% & Return the latest value when a fact changed across sessions. \\
entity\_resolution & 15 & 7.1\%  & Detect genuine ambiguity and ask which entity is meant. \\
\bottomrule
\end{tabular}
\normalsize
\end{table}

\begin{table}[H]
\centering
\caption{Representative confusable entity pairs (disambiguation targets). All company names
are fictitious.}
\label{tab:pairs}
\small
\begin{tabular}{ll}
\toprule
\textbf{Pair} & \textbf{Shared token(s)} \\
\midrule
Zephyron Cloud Solutions $\leftrightarrow$ Zephyron Cloud Systems & zephyron, cloud \\
Marvex Digital Marketing $\leftrightarrow$ Marvex Digital Solutions & marvex, digital \\
Calibrix Health Systems $\leftrightarrow$ Calibrix Financial Group & calibrix \\
Duronto Financial Services $\leftrightarrow$ Calibrix Financial Group & financial \\
Tandola Retail Group $\leftrightarrow$ Tandola Retail Solutions & tandola, retail \\
\bottomrule
\end{tabular}
\normalsize
\end{table}

\clearpage
\bibliographystyle{plainnat}
\bibliography{references}

\newpage

\end{document}